\documentclass[sigconf]{acmart}

\AtBeginDocument{%
  \providecommand\BibTeX{{%
    \normalfont B\kern-0.5em{\scshape i\kern-0.25em b}\kern-0.8em\TeX}}}

\acmYear{2024}

\acmConference[WACAI]{Workshop Affects, Compagnons Artificiels et Interactions
}{12-14 juin 2024}{Bordeaux (France)}
\begin{document}

\title{Investigating the impact of 2D gesture representation on co-speech gesture generation}

\author{T\'eo Guichoux}
\affiliation{%
  \institution{Sorbonne Université\\ISIR \\ STMS lab IRCAM \\ CNRS}
  \city{Paris}
  \country{France}
  \postcode{11}
}
\email{teo.guichoux@isir.upmc.fr}
\orcid{?}

\author{Laure Soulier}
\affiliation{%
 \institution{Sorbonne Université \\ ISIR, CNRS}
 \city{Paris}
 \country{France}}
\email{laure.soulier@isir.upmc.fr}

\author{Nicolas Obin}
\affiliation{%
  \institution{STMS lab \\ IRCAM, CNRS, Sorbonne Université}
  \city{Paris}
  \country{France}
}

\author{Catherine Pelachaud}
\affiliation{%
  \institution{Sorbonne Université \\ ISIR, CNRS}
  \city{Paris}
  \country{France}}
\email{catherine.pelachaud@isir.upmc.fr}

\renewcommand{\shortauthors}{T.Guichoux, L.Soulier, N.Obin, C.Pelachaud}

\begin{abstract}
  Co-speech gestures play a crucial role in the interactions between humans and embodied conversational agents (ECA). Recent deep learning methods enable the generation of realistic, natural co-speech gestures synchronized with speech, but such approaches require large amounts of training data. 
  "In-the-wild" datasets, which compile videos from sources such as YouTube through human pose detection models, offer a solution by providing 2D skeleton sequences that are paired with speech. Concurrently, innovative lifting models have emerged, capable of transforming these 2D pose sequences into their 3D counterparts, leading to large and diverse datasets of 3D gestures. However, the derived 3D pose estimation is essentially a pseudo-ground truth, with the actual ground truth being the 2D motion data. This distinction raises questions about the impact of gesture representation dimensionality on the quality of generated motions — a topic that, to our knowledge, remains largely unexplored. 
  In this work, we evaluate the impact of the dimensionality of the training data, 2D or 3D joint coordinates, on the performance of a multimodal speech-to-gesture deep generative model. We use a lifting model to convert 2D-generated sequences of body poses to 3D. Then, we compare the sequence of gestures generated directly in 3D to the gestures generated in 2D and lifted to 3D as post-processing. 
\end{abstract}



\received{8th March, 2024}

\maketitle

\section{Introduction}
\label{sec:intro}
\begin{figure*}
  \centering
  \includegraphics[width=\linewidth]{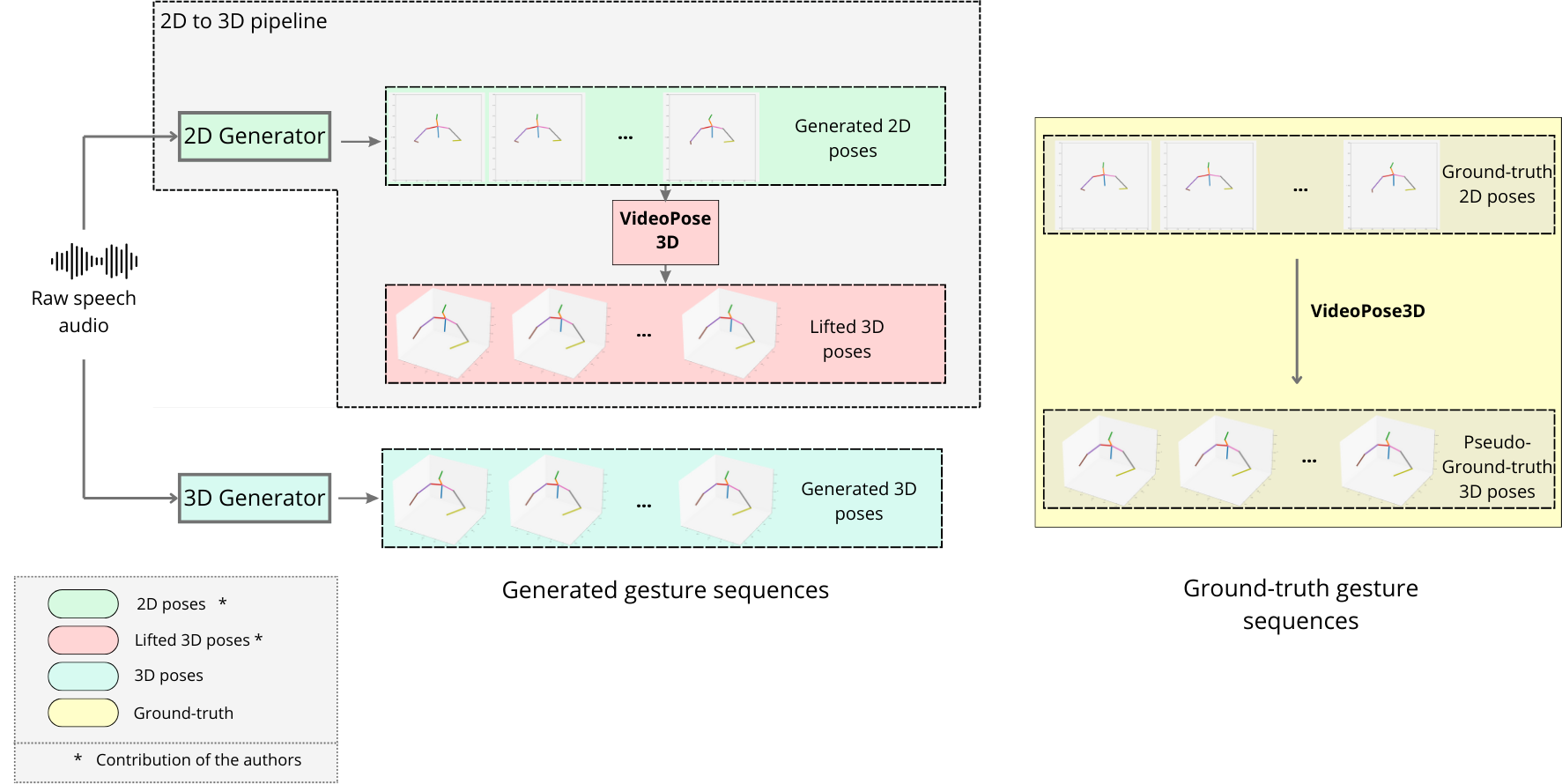}
  \caption{The proposed evaluation pipeline is a combination of DiffGesture \cite{zhu2023taming} that generates sequences of 2D body poses and VideoPose3D \cite{pavllo:videopose3d:2019} that lifts the generated 2D poses to 3D. The pseudo-ground-truth 3D gesture sequences originate from the TED Gesture-3D dataset \cite{Yoon2020Speech} and were obtained using VideoPose3D to lift 2D keypoints to 3D. The 2D keypoints were estimated using OpenPose \cite{cao2019Openpose} on TED YouTube videos.}
  \label{fig:pipeline}
\end{figure*}
In human communication, gestures play an integral role by conveying intentions and emphasizing points \cite{mcneill1994hand}.  Recent studies \cite{Alexanderson_2023, yang2023diffusestylegesture, yang2023qpgesture, Ao2023GestureDiffuCLIP, zhu2023taming, alexanderson2020style, Mehta_2023, voss2023aq, deichler2023diffusion, Yoon2020Speech, fares2023zero, yoon2018robot} aim to create similar gestures for Embodied Conversational Agents (ECA) to make interactions with humans more natural and effective. These new methods use learning algorithms and extensive human motion datasets to generate gestures alongside speech. The representation of co-speech gestures, in 2D or 3D, influences how the agent's non-verbal communication is perceived, especially the speaker's communication style \cite{habibie2021learning}.
Most recent work on co-speech gesture synthesis considers 3D motion data \cite{Alexanderson_2023, yang2023diffusestylegesture, yang2023qpgesture, Ao2023GestureDiffuCLIP, zhu2023taming, alexanderson2020style, Mehta_2023, voss2023aq, deichler2023diffusion, Yoon2020Speech}, primarily because such data representation is more expressive and more easily transferable to downstream applications such as animation, virtual reality or social robots \cite{Nyatsanga_2023, yoon2018robot}. However, the quality of learning-based co-speech gesture synthesis in terms of naturalness, speech synchrony, and diversity heavily relies on the quantity and quality of the training data which may be costly to collect.
In recent works, mostly two kinds of datasets are considered: motion-capture (Mocap) \cite{liu2022beat, ferstl2018trinity, ghorbani2022zeroeggs, lee2019talking, frestl2021trinity2} and pose estimation from "in-the-wild" videos \cite{yoon2018robot, voss2023aq, ahuja2020no, ginosar2019learning, habibie2021learning} which are videos freely accessible online.

Mocap datasets show obvious superiority in terms of fine-grained motion and annotation quality over "in-the-wild" datasets. However, they are very costly to collect and lack \textbf{1) }diversity because of the reduced number of considered speakers and \textbf{2)} naturalness because gestures and expressed emotions are acted in a controlled studio environment. Datasets based on "in-the-wild" videos, on the contrary, usually gather data from many different speakers with different speech and gesture profiles. They offer the possibility to access large raw datasets of online videos but with heterogeneous and uncontrolled situations and environments. Such datasets require robust and accurate pose estimation algorithms to estimate 2D keypoints coordinates from monocular videos \cite{cao2019Openpose, fang2017rmpe}. 3D body poses can further be inferred from these estimated 2D key-points using a third-party 2D to 3D lifting model \cite{pavllo:videopose3d:2019, ExPose:2020, mehta2019XNect}. Nevertheless, relying on such 3D lifters induces error inherently dependent on the ambiguous nature of 2D to 3D lifting problems, as one 2D pose can correspond to multiple 3D poses. 

Due to the ambiguous nature of 3D pose estimation from 2D keypoints, datasets leveraging 2D to 3D lifting models are prone to inaccuracies and lack of fine-grained motions such as hand joints and finger positions. 
Recent works on co-speech gesture synthesis from "in-the-wild" datasets mostly consider 3D data lifted from estimated 2D keypoints \cite{zhu2023taming, voss2023aq, liu2022learning, Yoon2020Speech, habibie2021learning}, but it remains unclear what is the impact of the dimensionality of the skeleton representation on the training of the generative model and the quality of the generated gestures.

The influence of the dimensionality of the gesture representation is clearly under-studied. Since "in-the-wild" datasets all use a 2D representation before estimating the 3D poses, we believe that the question of the data representation is fundamentally important. 

In this work, we compare two training settings to evaluate the influence of data dimensionality on the performance of a speech-to-gesture generative model by either considering 2D or 3D joint coordinates. We chose to evaluate the effect of the pose representation on a Denoising Diffusion Probabilistic Model (DDPM) because such models have proven their ability to generate natural and diverse gestures aligned with speech and are widely used in the co-speech gesture synthesis field \cite{zhu2023taming, Alexanderson_2023, Ao2023GestureDiffuCLIP, yang2023diffusestylegesture, deichler2023diffusion, zhao2023diffugesture}.
There is a one-to-many relationship between 2D keypoints and their 3D counterparts. Given the deterministic nature of the 2D to 3D lifter, it will consistently map any given 2D pose to the same corresponding 3D pose introducing an inductive bias in the process. We, therefore, formulate the following hypotheses:
\begin{itemize}
    \item \textbf{H1)} The distribution of lifted gestures cannot perfectly match the original 3D gesture distribution.
    \item \textbf{H2)} There is a drop in the consistency between speech and gestures when generating 2D gestures and lifting them to 3D.
    \item \textbf{H3)} The gestures generated in 2D and subsequently lifted to 3D are less diverse than those directly generated in 3D.
\end{itemize}
To verify these hypotheses, we propose the following contributions:
\begin{itemize}
    \item We propose an evaluation pipeline to investigate the impact of the dimensionality of the pose representation on the performance of a DDPM \cite{ho2020diffusion, sohl2015deep, song2019generative} for gesture generation. We train a speech-to-gesture DDPM \cite{zhu2023taming} to generate sequences of body poses represented in 2D coordinates which are then lifted to 3D using VideoPose3D \cite{pavllo:videopose3d:2019}. The pipeline is described in Figure~\ref{fig:pipeline}.
    \item We empirically compare the quality of the gestures generated in 2D lifted to 3D to the gestures directly generated in 3D using evaluation metrics commonly used in co-speech gesture generation tasks \cite{Yoon2020Speech, zhu2023taming, liu2022learning}. Specifically, we conduct a series of experiments using a 2D dataset from "in-the-wild" videos and its 3D counterparts.
\end{itemize}

The remainder of this paper is organized as follows: first, we present state-of-the-art works focusing on co-speech gesture generation. We then introduce our methodology and experimental design. Then, we discuss the experimental results and end with a conclusion.

\begin{table*}
\caption{Speech-Gesture datasets since 2019. The collection methods are described in the rightmost column. It can be either Motion Capture (MoCap) or pose estimation. Abbreviations: \textit{up.} \textit{upper}, \textit{rot.} \textit{rotations} \textit{coord.} \textit{coordinates}, \textit{n.s} \textit{not specified}. This list is an update of the one provided by Nyatsanga et al.\cite{Nyatsanga_2023}.}
\label{tab:dataset}
\small
\begin{tabular}{|l|l|l|l|l|l|}
\hline
Dataset &
  Size &
  \# of speakers &
  Type of motion data &
  Finger motion &
  Collection method
  \\ \hline
TED Gesture \cite{yoon2018robot} &
    52.7 h &
    1,295 &
    2D joint coord. &
    &
    OpenPose \cite{cao2019Openpose}  \\
    
TED Gesture 3D \cite{Yoon2020Speech} &
  97h &
  n.s. &
  3D joint coord. &
  &
  OpenPose \cite{cao2019Openpose}, 
  VideoPose3D \cite{pavllo:videopose3d:2019} \\

BiGe \cite{voss2023aq} &
  260h &
  n.s. &
  3D joint coord. &
  Yes &
   OpenPose \cite{cao2019Openpose},
   VideoPose3D\cite{pavllo:videopose3d:2019}  \\
  
TED Expressive \cite{liu2022learning} &
  100.8h &
  n.s. &
  3D joint coord. &
  Yes &
  OpenPose \cite{cao2019Openpose}, ExPose \cite{ExPose:2020} \\
  
PATS \cite{ahuja2020no} &
  250h &
  25 &
  2D joint coord. &
  &
  OpenPose \cite{cao2019Openpose}  \\

SpeechGesture \cite{ginosar2019learning} &
  144 h &
  10 &
  2D joint coord. &
  Yes &
  OpenPose \cite{cao2019Openpose} \\

SpeechGesture 3D \cite{habibie2021learning} &
  33h &
  6 &
  3D joint coord. &
  Yes &
  OpenPose \cite{cao2019Openpose}, XNect \cite{mehta2019XNect}, \cite{garrido2016reconstruction}\\
  
BEAT \cite{liu2022beat} &
  76h &
  30 &
  3D joint rot. &
  Yes &
  MoCap  \\
  
Trinity Speech Gesture I \cite{ferstl2018trinity}&
  6h &
  1 &
  3D joint rot. &
  &
  MoCap  \\

Trinity Speech Gesture II \cite{frestl2021trinity2}&
  4h &
  1 &
  3D joint rot. &
  &
  Mocap \\
  
ZeroEggs \cite{ghorbani2022zeroeggs}&
  2h &
  1 &
  3D joint rot. &
  Yes &
  MoCap \\

TalkingWithHands \cite{lee2019talking}&
    50h &
    50 &
    3D joint rot. &
    Yes &
    Mocap \\
   
\hline
\end{tabular}
\end{table*}

\section{Related Work}

\noindent \textbf{Learning-based co-speech gesture generation. } The co-speech gesture synthesis field has seen an important shift to deep learning approaches for gesture generation due to their effectiveness in creating natural movements that are well-synchronized with speech, with minimal assumptions \cite{Nyatsanga_2023}. 

Deterministic approaches that directly translate speech to gesture sequences have been proposed. To this end, one can choose different neural network architectures such as multi-layer perceptrons \cite{kucherenko2020gesticulator}, convolutional neural networks \cite{habibie2021learning}, recurrent neural network \cite{Yoon2020Speech, bhattacharya2021speech, liu2022learning, yoon2018robot} or transformers \cite{bhattacharya2021speech, windle2023the}.  Yoon et al. \cite{yoon2018robot} proposed a sequence-to-sequence model that was trained to generate 2D gesture sequences from the TED Gesture dataset. The generated pose sequences were then lifted to 3D to be mapped onto a social robot.

In recent works, there is a notable interest in non-deterministic generative models such as Variational Autoencoders (VAEs) \cite{li2021audio} and diffusion models \cite{ho2020diffusion,  song2019generative, Alexanderson_2023, deichler2023diffusion, Ao2023GestureDiffuCLIP, chemburkar2023discrete, tonoli2023gesture, zhao2023diffugesture}  due to their capacity for producing a wide array of gestures.

Specifically, VAEs are designed to encode gestures into a continuous latent space and subsequently decode these latent representations into speech-conditioned movements \cite{li2021audio}.  Recently, the gesture generation field has particularly focused on Probabilistic Denoising Diffusion Models \cite{ho2020diffusion,  song2019generative, Alexanderson_2023, deichler2023diffusion, Ao2023GestureDiffuCLIP, chemburkar2023discrete, tonoli2023gesture, zhao2023diffugesture} due to their capacity to robustly produce diverse and realistic gestures under multiple conditions, including speech, text, speaker identity, and style. In diffusion-based methods audio-driven gesture synthesis is generally executed through classifier-free guidance \cite{ho2022classifierfree, Alexanderson_2023, zhu2023taming, Ao2023GestureDiffuCLIP}, leveraging both conditional and unconditional generation mechanisms during the sampling process.
Alexanderson et al. \cite{Alexanderson_2023} used Conformers \cite{gulati2020conformer} to generate gestures conditioned on behavior style and speech audio.
Ao et al. \cite{Ao2023GestureDiffuCLIP} leverage CLIP \cite{clip2021radford} to encode speech text and a style prompt and use a combination of AdaIN \cite{huang2017adain} and classifier-free guidance to generate diverse yet style-conditioned gestures from speech.
Zhu et al. \cite{zhu2023taming} proposed DiffGesture, using a Diffusion Audio-Gesture Transformer to guarantee temporally aligned generation. In their work, raw speech audio is concatenated to gesture frames to condition the diffusion process. DiffGesture was trained on the TED Gesture-3D dataset \cite{Yoon2020Speech}, which compiles 3D gestures inferred from 2D poses obtained from monocular video. We hence used their model as a baseline for our study.
\\

\noindent \textbf{Representation and collection of the gesture data.} 
The quality and diversity of the training data are crucial for training co-speech gesture generative models. Early works mostly considered 2D motion data \cite{fares2023zero, yoon2018robot, ginosar2019learning}. 2D gestures were typically extracted from "in-the-wild" monocular videos using a third-party pose extractor such as OpenPose \cite{cao2019Openpose, yoon2018robot, ahuja2020no, ginosar2019learning}. See Table \ref{tab:dataset} for a list of existing gesture datasets. This collection process allows the gathering of a large amount of training data with numerous different speakers and ensures the diversity and spontaneity of the gestures. However, leveraging such pre-trained pose estimators induces errors resulting in low motion quality, especially for fine-grained motion such as fingers, and limits the pose representation to be two-dimensional. Most of the recent literature \cite{Alexanderson_2023, yang2023diffusestylegesture, yang2023qpgesture, Ao2023GestureDiffuCLIP, liu2022beat, zhao2023diffugesture, kucherenko2023genea} focuses on MoCap datasets \cite{ferstl2018trinity, frestl2021trinity2, liu2022beat, ghorbani2022zeroeggs, lee2019talking}. MoCap datasets capture detailed 3D movements in a studio, ensuring high-quality motion capture, including detailed finger movements and precise full-body keypoint positions. However, the limited number of speakers in the dataset and the controlled studio environment for data capture diminish the diversity and spontaneity of the training data, consequently affecting the variety of the gestures generated by models trained on such datasets \cite{Nyatsanga_2023, Alexanderson_2023, yang2023diffusestylegesture, yang2023qpgesture, Ao2023GestureDiffuCLIP, liu2022beat, zhao2023diffugesture, kucherenko2023genea}. Recently, multiple works \cite{zhu2023taming, Yoon2020Speech, voss2023aq, liu2022learning, habibie2021learning} opt for increased diversity and volume of data samples while keeping a 3D representation of gestures, choosing to train their models on datasets of 3D gestures collected from "in-the-wild" videos \cite{voss2023aq, Yoon2020Speech, zhu2023taming, habibie2021learning, liu2022learning}. To extract 3D body poses from monocular videos, the data collection process typically leverages a pipeline of pose extraction \cite{cao2019Openpose} and 2D-to-3D lifting \cite{pavllo:videopose3d:2019, ExPose:2020, mehta2019XNect}. For instance, the dataset TED Gesture-3D introduced by Yoon et al. \cite{Yoon2020Speech} leverages VideoPose3D \cite{pavllo:videopose3d:2019} to convert 2D body keypoints extracted by OpenPose \cite{cao2019Openpose} to 3D. This dataset is an extension of the previous TED Gesture dataset \cite{yoon2018robot} where the pose where represented in 2D. \footnote{To avoid confusion we refer to the 3D version of \cite{Yoon2020Speech} as \textit{TED Gesture-3D}}
\\

In this work, we study how training an audio-driven diffusion model to generate 2D motion data and then post-processing the generated sequences using a 3D lifter impacts the overall quality of the synthesized gestures. Specifically, we use DiffGesture \cite{zhu2023taming} as a gesture generator which obtained state-of-the-art results on the TED Gesture-3D and TED Expressive datasets \cite{Yoon2020Speech, liu2022learning}. For the 2D to 3D lifting model, we employ VideoPose3D \cite{pavllo:videopose3d:2019}. We use the TED Gesture 3D dataset \cite{Yoon2020Speech} for our evaluation, which is a dataset of 3D gestures extracted from YouTube videos.
We aim to highlight the presence of an inductive bias due to using a deterministic 2D-to-3D lifting model, and we evaluate its impact on the generated gestures' diversity, naturalness, and synchrony.

\section{Methodology}

The objective of this paper is to analyze how the dimensionality of body pose data influences the behavior of the generation of co-speech gestures. In this section, we present our evaluation pipeline, the model and the dataset we used as well as the chosen evaluation metrics.

\subsection{Data}

Our analysis is based on the TED Gesture-3D dataset \cite{Yoon2020Speech}. TED Gesture-3D is a dataset including pose sequences extracted from in-the-wild videos of TED talkers with the corresponding speaker identity, speech, and speech transcription. TED Gesture-3D includes 3D body poses estimated via a combination of a 2D pose extractor from monocular videos \cite{cao2019Openpose} and VideoPose3D \cite{pavllo:videopose3d:2019}.
The size of the dataset is 97h where the poses are sampled at 15 frames per second with a stride of 10 with a total of 252,109 sequences of 34 frames. 
Body poses are represented as vectors in $\mathcal{R}^{N\times J \times 3}$ where $N$ is the sequence length and $J$ is the number of body joints. Instead of considering raw joint coordinates for body pose representation, we follow the approach proposed by Yoon et al. \cite{Yoon2020Speech} where a body pose is represented as nine directional vectors where each direction represents a bone. The vectors are normalized to the unit length and centered on the root joint. This pose representation is invariant to bone length and less affected by root rotations therefore favoring the training.
In this work, 2D pose sequences are vectors of 3D poses from which the depth axis has been removed. 

\subsection{Pipeline}
\label{pipeline}
To evaluate the inductive bias caused by the dimensionality of the gesture representation (2D or 3D) and the 2D-to-3D conversion, we trained a co-speech gesture generator on both 2D and 3D settings and employed a 3D lifter for post-processing the 2D generated sequences to be able to compare them to the 3D generated sequences. The complete pipeline is described in Figure \ref{fig:pipeline}. To better understand the influence of the motion dimensionality on the relationship between speech and gesture we also perform an ablation on the gesture generator where the speech condition is removed.

\subsubsection{Gesture generator}

The co-speech gesture generator used as a reference in this study is defined as a DDPM which generates sequences of poses out of noise, conditioned on raw speech audio. DDPMs rely on two Markov chains: the forward process that gradually adds noise to the data and the backward process that converts noise to data. The backward process is modeled as a deep neural network that synthesizes gestures conditioned on speech. Raw audio is encoded using a convolutional neural network and then concatenated to the noisy pose sequence along the features axis.
We used DiffGesture proposed by Zhu et al \cite{ho2020diffusion, zhu2023taming} trained on the TED Gesture-3D dataset \cite{Yoon2020Speech}. The body poses are represented in $\mathcal{R}^{J\times3}$ where $J$ is the number of considered body joints. To synthesize diverse and speech-accurate gestures, DiffGesture uses classifier-free guidance \cite{ho2022classifierfree}. This approach involves jointly training a conditioned and an unconditioned DDPM, allowing for a trade-off between the quality and diversity of the generated poses at inference time.

DiffGesture was first designed to generate 3D gestures, we straightforwardly adapted the architecture to account for 2D body pose sequences by changing the input and output dimensions of the denoising network. Specifically, we removed the depth axis of the body pose coordinates thus considering poses in $\mathcal{R}^{J\times2}$, we refer to this version as \textit{DiffGesture 2D} and the original version is referred as \textit{DiffGesture 3D}.
We trained DiffGesture in four different settings: 2D motion generation, and 3D motion generation, with and without speech condition. For unconditional generation, we simply masked the speech input, forcing DiffGesture to directly generate gestures out of noise, without further guidance.  We obtained similar results as Zhu et al. \cite{zhu2023taming} when retraining \textit{DiffGesture 3D} demonstrating the validity of our evaluation protocol (see Table \ref{tab:experimental-results}). 

\subsubsection{2D-3D Lifter}
We employed a 2D-3D lifter defined by a temporal convolutional network (TCN). Specifically, we used VideoPose3D \cite{pavllo:videopose3d:2019} to lift 2D pose sequences to 3D. The lifting process is defined as a mapping problem, in which the TCN employs 1-D convolutions along the temporal axis to transform 2D full body poses into a temporally consistent sequence of 3D body poses. VideoPose3D utilizes dilated temporal convolutions to capture long-term information.

We retrained VideoPose3D \cite{pavllo:videopose3d:2019} on the TED Gesture-3D dataset to be able to input body poses in $\mathcal{R}^{2\times 9}$ i.e when only the upper part of the body is considered. We obtained a slightly better mean per joint positional error (MPJPE) when the sequences were up-scaled to 273 frames per second (fps) to exceed the receptive field of VideoPose3D instead of the original 15 fps. The final MPJPE of VideoPose3D was $\boldsymbol{11.1}$ on the test set of TED Gesture-3D. We kept the model architecture and training hyper-parameters consistent with the original implementation, except for the learning rate decay, which we found did not enhance the training process.

\subsection{Experimental set-up}

\subsubsection{Comparative settings}
\label{sec:experiments}
We aim to study the impact of the dimensionality of the motion data on the performance of a diffusion-based generative model. To this end, we considered three experimental settings. 

\textbf{1)} We want to evaluate the impact of training on 2D motions on the quality of 3D gesture sequences. For this experiment, we define \textit{DiffGesture 2D + VP3D} as DiffGesture trained on 2D motion data whose outputs are then lifted to 3D using VideoPose3D and we compare it to the original DiffGesture \cite{zhu2023taming}.

\textbf{2)} To further explore the impact of motion dimensionality on the generated gesture, we also compare \textit{DiffGesture 2D} to DiffGesture but where the 3D generated motion is narrowed to 2D by removing the depth axis, we refer to this model as \textit{DiffGesture 3D->2D}.

\textbf{3)} To evaluate the effect of motion dimensionality on multimodality we proceeded to an ablation of DiffGesture where the input speech is masked during training and inference. In practice, this is equivalent to the original conditional training of DiffGesture with a masking probability $p_{uncond}$ always set to $1$. We performed these experiments in both 2D and 3D settings. We refer to these models as \textit{Uncond. DiffGesture 2D} and \textit{Uncond. DiffGesture 3D}. 

\subsubsection{Evaluation metrics}
\label{sec:metrics}

We empirically evaluate our models with three commonly used metrics in the co-speech gesture generation field. 

The \textbf{Fréchet Gesture Distance} (FGD) defined by Yoon et al. \cite{Yoon2020Speech} is an adaptation of the Fréchet Inception Distance (FID) \cite{gans2017Heusel}. The FGD computes the 2-Wasserstein distance between two distributions leveraging latent features extracted with a pose encoder. Similar distributions will result in a high FGD value. The FGD is defined as follows:
\begin{equation}
    FGD(X,\hat{X}) = \lVert \mu_r - \mu_g \rVert + Tr(\Sigma_r + \Sigma_g - 2(\Sigma_r \Sigma_g)^2 )
\end{equation} 
Where: $X$, $\hat{X}$ are the real and generated distributions respectively; $\mu_r$, $\mu_g$, $\Sigma_r$, $\Sigma_g$ are the mean and covariance of the latent distributions extracted from the real and generated distributions.

The \textbf{Beat Consistency Score} (BC) measures the temporal consistency between kinematic and audio beats of a paired audio-motion sequence. This measure first introduced for evaluating the synchrony of dance with music \cite{dance2021Li}, has been adapted to speech and gestures \cite{learn2021Li}.
First, kinematic beats are extracted from a pose sequence by selecting the time steps of the sequence where the average angle velocity is higher than a certain threshold. Angle velocity is computed using the variation in angle between two successive frames. Intuitively, BC measures the average distance between the time steps corresponding to an audio beat and the closest time steps corresponding to a kinematic beat. The audio beats are extracted using a pre-trained detection model and the BC score is computed as follows:
\begin{equation}
\label{eq:bc}
     BC = \frac{1}{n}\sum_{i=0}^n exp \Big ( - \frac{min_{\forall t_{j}^y \in \mathcal{B}^y} \lVert t_{j}^y - t_{i}^x \rVert ^2}{2\sigma^2} \Big )
\end{equation}
Where: $t_{i}^x$ is the $i-th$ audio beats, $B_y = {t_{i}^y }$ is the set of the kinematic beats of the $i-th$ sequence, and $\sigma$ is a parameter to normalize sequences, set to 0.1 empirically as in \cite{zhu2023taming}. 

The \textbf{Diversity} measure also leverages the latent features extracted with a pose encoder \cite{hsin2019diversity}. Diversity is computed by randomly selecting two sets of $N$ features from the generated distribution and calculating the distance between the mean of both sets in the feature space.  Typically, if a model generates similar gestures all gestures will be close to the average gesture sequence, resulting in a small distance between the two sets, as formalized below:
\begin{equation}
\label{eq:diversity}
    Div(X) = \lVert \mu_{A} - \mu_{B} \rVert_2 
\end{equation}
Where: $X$ is a distribution of gestures, $A$ and $B$ are sets of gestures randomly sampled from $X$ and $\mu_A$ and $\mu_B$ are the mean of the gesture features in both sets.

\section{Results and Discussion}

\begin{table}
\caption{Experimental results of experiments on the TED Gesture-3D dataset \cite{Yoon2020Speech}. These results correspond to the experiments (1) and (2) in section \ref{sec:experiments}.
Up arrows indicate that a higher result is better whereas down arrows indicate that a lower result is better. * means reported results from \cite{zhu2023taming}}
\label{tab:experimental-results}
\begin{tabular}{lccc}
\hline
\multicolumn{1}{c}{} &
  \multicolumn{3}{c}{TED Gesture }  \\ \cline{2-4} 
Methods &
  FGD $\downarrow$ &
  BC $\uparrow$ &
  Diversity $\uparrow$ 
  \\
\hline
\hline
Evaluation on the 3D gesture space & & \\
\hline

\textbf{Ground Truth 3D} &
    0 &
    0.702 &
    102.339  \\
DiffGesture 3D \cite{zhu2023taming} &
  1.370 &
  0.659 &
  102.586  \\
DiffGesture 2D + VP3D &
  9.833 &
  0.571 &
  92.136  \\

\hline
\hline
Evaluation on the 2D gesture space & & \\
\hline

\textbf{Ground Truth 2D} &
  0 &
  0.689 &
  112.76 \\

DiffGesture (3D->2D) &
  1.722 &
  0.645 &
  110.649  \\
  
DiffGesture 2D &
  3.279 &
  0.643 &
  112.165 \\

\hline
\hline
Reported results from \cite{zhu2023taming}  & &\\
\hline

\textbf{Ground Truth 3D} &
    0 &
    0.698 &
    108.525  \\
DiffGesture * \cite{zhu2023taming} &
  1.506 &
  0.699 &
  106.722 \\
Attention Seq2Seq* \cite{yoon2018robot} &
  18.154 &
  0.196 &
  82.776  \\
Speech2Gesture* \cite{ginosar2019learning}&
  19.254 &
  0.668 &
  93.802  \\
Joint Embedding* \cite{ahuja2019language}&
  22.083 &
  0.200 &
  90.138  \\
Trimodal* \cite{Yoon2020Speech}&
  3.729 &
  0.667 &
  101.247 \\
HA2G* \cite{liu2022learning}&
  3.072 &
  0.672 &
  104.322 \\

\hline
\end{tabular}
\end{table}

\begin{table}
\caption{Ablation study of DiffGesture \cite{zhu2023taming} on the TED Gesture-3D dataset \cite{Yoon2020Speech} where the speech condition has been removed. Up arrows indicate that a higher result is better whereas down arrows indicate that a lower result is better.}
\label{tab:ablation}
\begin{tabular}{lccc}
\hline
\multicolumn{1}{c}{} &
  \multicolumn{3}{c}{TED Gesture }  \\ \cline{2-4} 
Methods &
  FGD $\downarrow$ &
  BC $\uparrow$ &
  Diversity $\uparrow$ 
  \\
\hline
\hline
Evaluation on the 3D gesture space & & \\
\hline

\textbf{Ground Truth 3D} &
    0 &
    0.702 &
    102.339  \\
Uncond. DiffGesture 3D &
  3.288 &
  0.683 &
  98.905  \\
Uncond. Diff Gesture 2D + VP3D &
  10.009 &
  0.595 &
  93.945  \\

\hline
\hline
Evaluation on the 2D gesture space & & \\
\hline

\textbf{Ground Truth 2D} &
  0 &
  0.689 &
  112.76 \\

Uncond. DiffGesture (3D->2D)  &
  5.529 &
  0.667 &
  111.599  \\
  
Uncond. DiffGesture 2D  &
  1.757 &
  0.653 &
  113.304  \\

\hline
\end{tabular}
\end{table}

The results of our experiments are reported in Table \ref{tab:experimental-results}. In the table's upper section, we present outcomes from our experiments evaluating gestures in 3D. The middle section details the results from our experiments assessing gestures in 2D. The results from Zhu et al. \cite{zhu2023taming} are reported in the table's lower section. It is important to note that we retrained the motion encoder used to compute the FGD and diversity score. The reported results from Zhu et al. were obtained using their own encoder.
We reported the results of our ablation study in Table \ref{tab:ablation}.
\\

\noindent \textbf{Evaluation of lifted generated gestures.} When comparing the results of \textit{DiffGesture 2D + VP3D} to those of \textit{DiffGesture 3D}, we can notice that \textit{DiffGesture 2D + VP3D} performs worse than the original \textit{DiffGesture 3D} in terms of FGD, BC, and diversity. The drop in FGD validates the \textbf{H1} hypothesis as the FGD measures the distance between two distributions of gestures. Similarly, the drop in diversity confirms the hypothesis \textbf{H3}. We assume that the one-to-many relationship between 2D and 3D keypoints is mostly responsible for the performance drop of \textit{DiffGesture 2D + VP3D} for the FGD and diversity. As VideoPose3D is deterministic, to one 2D pose it will systematically predict the same 3D pose although there exists multiple possibilities. Hence, the distribution resulting from lifting 2D sequences is tighter than the distribution directly generated in 3D, explaining the high FGD and low diversity of the gestures generated in \textit{DiffGesture 2D + VP3D}.
There is a drop in BC between \textit{DiffGesture 3D} and \textit{DiffGesture 2D + VP3D} which corroborates the \textbf{H2} hypothesis. We think that post-processing 2D gestures using VideoPose3D tends to over-smooth the resulting 3D gestures, reducing the number of kinematic beats.
\\

\noindent \textbf{Evaluation of the quality of gestures generated in 2D.} When evaluating in the 2D motion space, \textit{DiffGesture 3D->2D} performs better than \textit{DiffGesture 2D} in terms of FGD. Hence, training DiffGesture to generate 3D motion sequences seems to behave better than training the model on 2D motion data. This outcome was anticipated since the representation of poses in 3D is more detailed compared to the 2D version. 
The BC and diversity scores do not seem to be influenced by the dimensionality of the gestures that were used to train the generative model.
\\

\noindent \textbf{Ablation of the speech condition.}
The results of the ablation study are reported in Table \ref{tab:ablation}.
\noindent \textit{Uncond. DiffGesture 3D} shows worse FGD and diversity than \textit{DiffGesture 3D}. This outcome demonstrates that the speech condition helps \textit{DiffGesture 3D} to synthesize gesture sequences that are both diverse and similar to the target distribution.
As depicted in Table \ref{tab:experimental-results}, \textit{DiffGesture 2D} shows a higher FGD value than \textit{DiffGesture (3D->2D)} and \textit{Uncond. DiffGesture 2D}, suggesting that the speech condition reduces the quality of the gestures generated in 2D. We are led to think that the integration of a speech condition adds ambiguity to the generation of 2D gestures. A 2D gesture can correspond to multiple 3D gestures, and there is a one-to-many relationship between 3D gestures and speech. Therefore reducing the gesture dimension may lead to a more pronounced one-to-many relationship between gestures and speech. 
Such findings suggest that incorporating a speech condition enhances the diversity of gestures generated by \textit{DiffGesture 3D}. In contrast, \textit{Uncond. DiffGesture 2D} manages to produce samples that maintain a diversity level comparable to that of the ground truth. This indicates that 2D gestures can be synthesized without the speech condition while still closely aligning with the target distribution in terms of FGD and diversity. This needs to be more studied with other models and other databases.  

It is important to note that the BC score is almost unaffected when removing the speech condition as depicted in Table \ref{tab:ablation}. More experiments involving other datasets and other models are needed to verify this behavior and are left for future work.
\\

Our objective evaluation validated our three hypotheses \textbf{H1}, \textbf{H2} and \textbf{H3} (c.f section \ref{sec:intro}). We can conclude that generating gestures in 2D and subsequently lifting them to 3D significantly impairs the overall quality of the gestures in terms of FGD, BC, and diversity. We believe that the 2D-to-3D conversion is mainly responsible for the performance drop. However, as gestures generated in 2D show worse FGD than gestures generated in 3D and narrowed to 3D, we think that the 3D representation of gestures favors the training of the generative model. This also shows that, even when generating 2D gestures, it is better to train DiffGesture to generate 3D gestures and narrow them to 2D as post-processing.

\section{Conclusion}

In this study, we explored how training a diffusion-based co-speech gesture generator with 2D data affects its performance. We introduced a pipeline that pairs a gesture generator with a 2D-to-3D lifting model, specifically VideoPose3D \cite{pavllo:videopose3d:2019}. Our findings reveal that using this pipeline negatively impacts overall performance. Using a deterministic lifting model, such as VideoPose3D, reduces the diversity of the generated gestures. 3D gestures lifted from 2D generated gestures are also less similar to the target 3D gesture distribution in comparison to gestures generated directly in 3D.  However, using such a 2D-to-3D lifter remains a feasible strategy when retraining the model with 3D data (e.g. MoCap data) is not an option. We also found that DiffGesture faces challenges in accurately modeling the relationship between 2D motion data and speech, a problem that does not occur with 3D motion data. We attribute this issue to the inherent ambiguity of 2D coordinates compared to 3D, which hinders the model's ability to synchronize the two modalities effectively. However, further evaluations on other datasets are needed to confirm this tendency.
In the TED Gesture-3D dataset, the 3-dimensional gestures are lifted from 2D body poses. Our evaluation is therefore biased as we do not have access to the real 3D ground truth.  For future research, we plan to conduct a similar analysis using a Mocap dataset. Additionally, we aim to employ an alternative generative model to validate our findings.

\bibliographystyle{ACM-Reference-Format}
\bibliography{references}

\appendix

\end{document}